%% file: main.tex
\pdfoutput=1  
\documentclass[10pt,twocolumn,letterpaper]{article}

\usepackage[pagenumbers]{cvpr}   
\input{preamble}

\definecolor{cvprblue}{rgb}{0.21,0.49,0.74}
\usepackage[breaklinks,colorlinks,allcolors=cvprblue]{hyperref}

\def\confName{CVPR}
\def\confYear{2027}

\title{AutoThreshold: Joint Confusion-Profile Learning for Adaptive Mask Binarization}

\author{%
Shaokai Wu$^{1,2}$, \quad Xunchu Zhou$^{1,2}$, \quad Pinzhi Wang$^{2}$, \quad Bo Dong$^{2}$\\[3pt]
Qi Kong$^{2}$, \quad Xiao Cui$^{2}$, \quad Yue Ding$^{1}$, \quad Hongtao Lu$^{1}$\\[10pt]
{\small $^{1}$Shanghai Jiao Tong University \qquad $^{2}$Zelos Technology}}

\begin{document}
\maketitle

\input{sec/0_abstract}
\input{sec/1_intro}
\input{sec/2_related}
\input{sec/3_formulation}
\input{sec/4_method}
\input{sec/5_experiments}
\input{sec/6_conclusion}

{
\small
\bibliographystyle{ieeenat_fullname}
\bibliography{main}
}

\end{document}

%% file: preamble.tex
\newcommand{\Fb}{F_{\beta}}
\newcommand{\CE}{\mathrm{CE}}

%% file: sec/0_abstract.tex
\begin{abstract}
A single validation-tuned threshold can become brittle when dense-score distributions shift. We introduce AutoThreshold, a post-hoc policy that predicts the joint mass of positive and negative pixels over score bins. This target directly parameterizes threshold-indexed confusion counts, avoiding products of independently estimated prevalence and class-conditional rates. Softmax normalization and cumulative decoding yield coherent, monotone confusion profiles; a plug-in decoder then selects a threshold for any $\Fb$ objective without retraining the head. Across three independently trained Kvasir pipelines transferred to four polyp datasets, Joint exceeds continuous Otsu by $0.0080 \pm 0.0046$ external macro Dice, with positive gains in 3/3 runs. On the reference pipeline, three head seeds achieve $0.6324 \pm 0.0027$, 0.0085 above Otsu; direct-utility Ridge achieves 0.6342, with the paired difference interval spanning zero. The same Joint checkpoints outperform Otsu for $F_{0.5}$, Dice, and $F_2$. On developmental KITTI Road leave-one-corruption-family-out evaluation, Joint improves over scalar regression by 0.0082 [0.0064, 0.0101], while direct utility is 0.0022 higher. These results establish joint confusion prediction as a useful, metric-retargetable inductive bias, without claiming Bayes-optimal $\Fb$ decisions.
\end{abstract}

%% file: sec/1_intro.tex
\begin{figure*}[t]
  \centering
  \includegraphics[width=\textwidth]{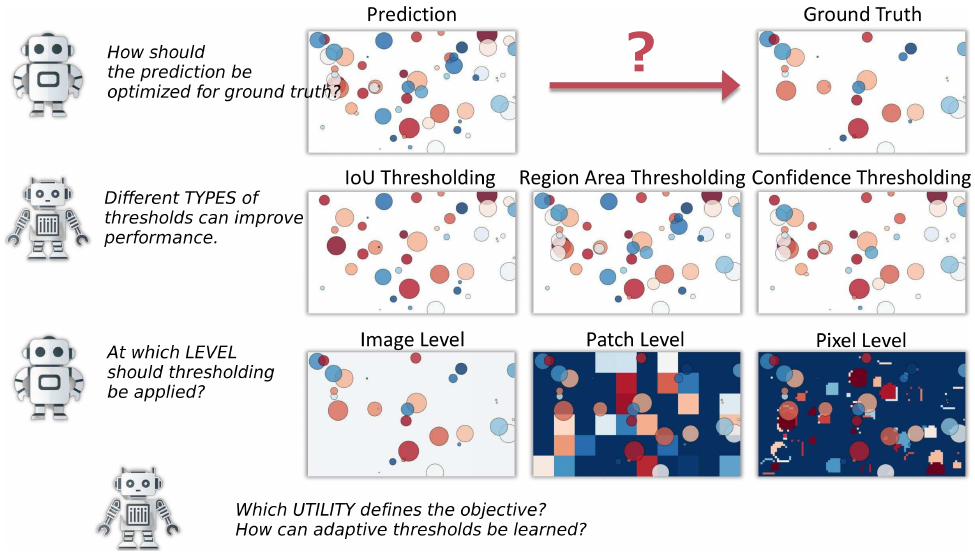}
  \caption{Design space for threshold policies. Scored outputs admit different threshold types, spatial supports (image, patch, or dense location), and decision-unit utilities. We instantiate image-level mask binarization; other supports are valid only when both operator and utility decompose accordingly. Color denotes confidence, and size illustrates a second thresholdable attribute.}
  \label{fig:design}
\end{figure*}

\section{Introduction}
\label{sec:intro}

Vision models produce continuous scores, but deployed systems ultimately act on discrete masks. The binarization threshold directly trades false positives against false negatives, yet it is commonly fixed once on source validation data. That choice ignores image- and domain-dependent changes in score calibration, score shape, and foreground prevalence---precisely the changes encountered after deployment.

Adaptive thresholding is well established in low-level image processing~\citep{roy2014adaptive}, and task-specific systems adapt training or post-processing thresholds in object detection and crowd localization~\citep{liu2019adaptive,wu2023boosting,yang2023cpss,gao2020learning}. Learned mask binarization, however, presents a specific supervision choice. Regressing one tie-broken optimum discards the utility landscape: two equally distant errors can incur very different regret. Regressing every sampled utility is a principled finite-action value estimator, but it imposes no structure across neighboring thresholds and must be retrained when the target metric changes.

We seek a post-hoc target between these two extremes: richer than a scalar optimum, structured across thresholds, and independent of a particular $\Fb$. Binary masks expose exactly such a target. The joint distribution of pixel label and score bin determines $TP(t)$, $FP(t)$, and $FN(t)$ at every threshold; these confusion profiles in turn support the full family of $\Fb$ plug-in decisions.

AutoThreshold predicts positive and negative score-bin mass as one normalized distribution. Cumulative decoding enforces monotone confusion-count curves, and a metric decoder selects a threshold for Dice, $F_{0.5}$, or $F_2$ from the same prediction. An auxiliary loss aligns its score marginal with the observed, label-free histogram. The policy remains fully detached from the segmenter: labels construct offline targets from cached score maps, whereas deployment uses only score-map statistics and a lightweight MLP.

Our main contributions are:
\begin{itemize}
\item We formulate adaptive mask binarization as joint confusion-profile prediction. The target avoids products of separately estimated prevalence and conditional rates, while its cumulative decoder supports monotone, retraining-free $\Fb$ retargeting.
\item We separate the benefit of the structured target from capacity and objective choices through scalar, hard-grid, soft-utility, expected-regret, direct value-regression, Ridge, GBRT, Otsu, and area-only controls.
\item We test transfer across four polyp datasets and six held-out KITTI corruption families, with complete-pipeline repeats, two external aggregation schemes, paired uncertainty, and sensitivity and profile-error analyses.
\end{itemize}

%% file: sec/2_related.tex
\section{Related Work}
\label{sec:related}

\noindent\textbf{Post-Processing Thresholds} High-level vision systems use confidence thresholds to filter predictions and IoU thresholds in both post-processing and evaluation. These roles differ: an NMS overlap threshold controls suppression among predicted boxes~\citep{bodla2017softnms}, whereas evaluation IoU thresholds determine whether detections match ground truth~\citep{lin2014coco}. Figure~\ref{fig:design} situates mask binarization in this broader design space. Our joint decoder specifically exploits binary score maps; it does not extend directly to NMS or corpus-level ranking.

\noindent\textbf{Operator-Specific Thresholding} Algorithmic methods encode a particular decision rule. Adaptive NMS changes suppression according to local box density~\citep{liu2019adaptive}; DIoU supplies a distance-aware NMS criterion, while the CIoU study incorporates geometric factors into Cluster-NMS~\citep{zheng2020distance,zheng2022enhancing}. Threshold-adaptive focal loss changes how target-domain samples contribute during training~\citep{yan2023threshold}, while classical image methods derive binarization thresholds from intensity statistics~\citep{roy2014adaptive,song2022two}. These methods exploit operator-specific structure but do not learn a confusion representation that can be decoded for multiple task metrics.

\noindent\textbf{Learned Threshold Policies} Learned threshold methods typically attach adaptation to one task: CPSS-FAT adapts positive-sample selection thresholds during detector training~\citep{yang2023cpss}, Independent Instance Maps learn image-dependent spatial binarization maps for crowd localization~\citep{gao2020learning}, and Crowd Hat and CTFNet learn region-adaptive NMS or confidence-foresight mechanisms~\citep{wu2023boosting,wu2022faster}. AutoThreshold instead learns a detached mask policy from cached outputs of a specified frozen operator.

\noindent\textbf{Contextual Decisions and Calibration} Post-hoc calibration aligns confidence with empirical correctness~\citep{guo2017calibration}, while selective prediction learns when to abstain at a desired coverage~\citep{geifman2017selective}. They address score reliability or risk--coverage behavior, not per-image mask thresholds for a task metric. Decision-focused learning trains predictions for downstream optimization value~\citep{donti2017task,wilder2019melding}. We share that motivation but predict threshold-sufficient confusion statistics from cached outputs, requiring neither a differentiable optimizer nor changes to the frozen model.

Given a finite threshold grid and measured rewards for every action, direct utility regression is a full-information cost-sensitive plug-in rule~\citep{elkan2001foundations}; contextual-bandit methods address the harder partial-feedback setting~\citep{beygelzimer2011contextual,dudik2011doubly}. We include the resulting decision-consistent finite-action estimator as our strongest learned baseline. AutoThreshold instead predicts a metric-independent joint distribution over labels and score bins, sharing cumulative structure across neighboring thresholds.

\noindent\textbf{Metric-Aware Binary Decisions} Optimal $F_1$ thresholding and plug-in rules for generalized classification metrics derive decisions from calibrated probabilities or estimated confusion quantities~\citep{lipton2014optimal,kotlowski2017surrogate}. We use the same dependence of $\Fb$ on confusion counts, but at a different decision unit: our head predicts an entire threshold-indexed confusion curve for each frozen score map. Because $\Fb$ is nonlinear, evaluating predicted mean confusion counts is not generally equivalent to maximizing conditional expected $\Fb$. We therefore study a structured plug-in rule, not a decision-consistency claim.

%% file: sec/3_formulation.tex
\section{Problem Formulation}
\label{sec:formulation}

\noindent\textbf{Problem Definition} Let $x$ denote a decision unit, such as an image or a spatial region. A frozen model produces raw output $Y_x = \mathcal{M}(I)_x$ and a label-free descriptor $F_x$. Given ground truth $g_x$, a thresholded operator $\mathcal{P}(Y_x,\tau,t)$, and bounded scalar utility $m(\mathcal{P}(Y_x,\tau,t),g_x)$, we learn a policy $\pi(F_x)$ for threshold type $\tau$. Its deployment objective is
\begin{equation}
J(\pi) = \mathbb{E}_x\!\left[m\!\left(\mathcal{P}\!\left(Y_x,\tau,\pi(F_x)\right),g_x\right)\right].
\label{eq:deployment}
\end{equation}
This expectation fixes the method's scope: $J$ must decompose over units receiving independent thresholds. Macro image Dice and mean regional utility satisfy this condition. A corpus-level ranking functional such as standard COCO AP does not, because changing one image can alter other images' ranking contributions.

\noindent\textbf{Data Protocol} We split labeled data into disjoint $\mathcal{D}_{\text{train}}$, $\mathcal{D}_{\text{val}}$, and $\mathcal{D}_{\text{test}}$. The frozen model runs once on each split. Utility labels from $\mathcal{D}_{\text{train}}$ fit the policy; $\mathcal{D}_{\text{val}}$ labels serve only checkpoint and hyperparameter selection. Test labels score the selected policy but never fit or select it.

\noindent\textbf{Overview} For each training image, AutoThreshold records the joint mass of positive and negative pixels in every score bin. A lightweight, detached predictor $N$ maps label-free score-map features $F_x$ to this target. At deployment, cumulative decoding reconstructs a confusion profile over $\mathcal{A} = \{\alpha_1,\dots,\alpha_k\}$, and a chosen $\Fb$ decoder returns its predicted maximizer. This step requires neither labels nor online utility evaluation.

We detach the threshold policy from the vision model for three reasons:
\begin{itemize}
\item It leaves the original architecture unchanged and avoids redesigning an adaptive head for every task-model pair.
\item The same cached score maps support direct decision baselines and multiple $\Fb$ decoders.
\item Offline utility annotation avoids backpropagating through discrete threshold decisions.
\end{itemize}
The small policy thus amortizes a discrete decision evaluated offline.

%% file: sec/4_method.tex
\section{Learning Adaptive Threshold Policies}
\label{sec:method}

\subsection{Threshold Policy Design}
\label{sec:policy_design}

Before learning a threshold policy, one must specify both its action and its evaluation unit. Figure~\ref{fig:design} organizes these choices along three axes.

\noindent\textbf{Threshold Type:} Thresholds control distinct post-processing decisions. For example, Crowd Hat adapts regional NMS thresholds~\citep{wu2023boosting}, whereas IIM predicts a dense binarization-threshold map for crowd localization~\citep{gao2020learning}.

\noindent\textbf{Decision Unit:} A policy may share one image-wide threshold, predict one threshold per region, or produce a dense threshold field. ``Pixel-level'' in Figure~\ref{fig:design} is the raster-output special case of the last setting. A globally coupled decision, such as suppression, cannot be localized unless its operator is regionalized as well.

\noindent\textbf{Decision Utility:} Different utilities induce different optimal actions. We therefore annotate the same quantity that Equation~\eqref{eq:deployment} averages at deployment. It is not interchangeable with a non-decomposable corpus metric, even when the two share a name or scale.

Our experiments fix these choices to one mask-binarization threshold per image, supervised and evaluated by image-level $\Fb$.

\subsection{Offline Utility Annotation}
\label{sec:annotation}

\noindent\textbf{Optimizing Thresholds} Because $\mathcal{M}$ is frozen, every threshold candidate reuses $Y_x$. We define the sampled utility profile and its deterministic scalar summary as
\begin{equation}
\begin{aligned}
u_j(x) &= m(\mathcal{P}(Y_x,\tau,\alpha_j),g_x),\\
t^{*}(x) &= \alpha_{r(\arg\max_j u_j(x))},
\end{aligned}
\label{eq:utility}
\end{equation}
where $r$ is a fixed tie rule. We retain these labels for utility, scalar, and grid-classification baselines. AutoThreshold instead uses the threshold-sufficient confusion statistics below. The same construction applies to a $p\times p$ regional policy only when both operator and utility decompose by region; coupled decisions require a joint decision unit.

\noindent\textbf{Annotating Threshold Labels} For each unit in $\mathcal{D}_{\text{train}}$ we evaluate a fixed uniform grid and cache $(F_x,t^{*}(x),\mathbf{u}(x))$ for the baselines. For the joint head, we cache normalized positive and negative pixel mass over score bins. A separate validation cache supports model selection, but its tuples never update $N$. Reported test data enter neither training nor selection.

\subsection{Joint Confusion-Profile Learning}
\label{sec:joint}

Only cached training tuples update $N$; neither the segmenter nor hard binarization participates in backpropagation.

\noindent\textbf{Threshold Head} A descriptor may be pooled from internal features~\citep{lin2014network} or computed from the frozen score map. We use the latter: quantiles, histograms, moments, entropy, and thresholded area ratios feed a small ReLU MLP~\citep{nair2010rectified}. For $B$ score bins, its $2B$ logits parameterize one normalized joint distribution.

\noindent\textbf{Joint Target and Decoder} For $N$ valid pixels with score $s_p$ and binary label $y_p$, define the positive and negative joint mass in score bin $b$ as
\begin{equation}
\begin{aligned}
a_b &= \frac{1}{N}\sum_p \mathbf{1}[y_p=1,\;s_p\in b],\\
b_b &= \frac{1}{N}\sum_p \mathbf{1}[y_p=0,\;s_p\in b].
\end{aligned}
\label{eq:jointmass}
\end{equation}
The observed, label-free score marginal is $q_b = a_b + b_b$. A single softmax maps the logits to $(\hat{\mathbf{a}},\hat{\mathbf{b}})\in\Delta^{2B-1}$. Let $w_{jb}\in[0,1]$ denote the fraction of bin $b$ above threshold $\alpha_j$, providing within-bin interpolation. We decode
\begin{equation}
\begin{aligned}
\widehat{TP}_j &= \sum_b w_{jb}\hat{a}_b,\quad \widehat{FP}_j = \sum_b w_{jb}\hat{b}_b,\\
\widehat{FN}_j &= \sum_b \hat{a}_b - \widehat{TP}_j.
\end{aligned}
\label{eq:decode}
\end{equation}

This target preserves prevalence--score dependence. In particular, it avoids substituting the generally unequal product $\mathbb{E}[\rho\mid F]\mathbb{E}[R^{+}\mid F]$ for $\mathbb{E}[\rho R^{+}\mid F]$. Nonnegativity and normalization keep the estimated confusion counts coherent, while cumulative decoding makes TP and FP non-increasing in the threshold. We regularize the predicted marginal $\hat{q}_b = \hat{a}_b + \hat{b}_b$ toward the current score histogram $q_b$, but do not impose equality.

\noindent\textbf{Joint Objective and Metric Retargeting} We jointly supervise the cumulative counts, joint density, and observed marginal:
\begin{equation}
\begin{aligned}
\mathcal{L}_{\text{joint}} ={}& \frac{1}{k}\sum_j \left[(\widehat{TP}_j-TP_j)^2 + (\widehat{FP}_j-FP_j)^2\right]\\
&+ \lambda_{ab}\,\CE([\mathbf{a};\mathbf{b}],[\hat{\mathbf{a}};\hat{\mathbf{b}}]) + \lambda_q\,\CE(\mathbf{q},\hat{\mathbf{q}}),
\end{aligned}
\label{eq:loss}
\end{equation}
where $\CE(p,q) = -\sum_i p_i\log q_i$ and $\lambda_{ab}=\lambda_q=.01$. One predicted profile then supports any sampled $\Fb$ decoder:
\begin{equation}
\begin{aligned}
\hat{U}_{\beta,j} &= \frac{(1+\beta^2)\widehat{TP}_j}{(1+\beta^2)\widehat{TP}_j + \beta^2\widehat{FN}_j + \widehat{FP}_j},\\
\hat{t}_{\beta} &= \alpha_{\arg\max_j \hat{U}_{\beta,j}}.
\end{aligned}
\label{eq:retarget}
\end{equation}
Changing the precision--recall preference therefore changes only the decoder, not the learned head.

This is deliberately a plug-in rule. Even perfect conditional mean confusion counts generally satisfy $\Fb(\mathbb{E}[C\mid F])\neq\mathbb{E}[\Fb(C)\mid F]$. Direct utility regression has a finite-grid decision-consistency property that our joint decoder does not. The empirical hypothesis is instead that cumulative sharing supplies a useful finite-sample inductive bias while enabling metric retargeting.

\noindent\textbf{Direct Decision Baselines} A scalar head regresses $t^{*}$ with SmoothL1, while hard classification predicts its grid index. We additionally compare soft utility classification, direct expected-regret optimization, and unconstrained utility regression:
\begin{equation}
\mathcal{L}_{\text{util}} = \frac{1}{k}\sum_j(\hat{u}_j-u_j)^2,\quad \hat{t} = \alpha_{\arg\max_j \hat{u}_j}.
\label{eq:utilloss}
\end{equation}
The last objective is standard finite-action value regression. Under squared loss, its population prediction is $\mathbb{E}[\mathbf{u}\mid F]$, whose largest component maximizes conditional expected sampled utility. This makes it a stringent comparator, but it provides neither cross-metric reuse nor our structured confusion representation.

\noindent\textbf{Sampled Oracle Performance} For a labeled evaluation set of $n$ decision units, annotation yields
\begin{equation}
J_{\text{oracle}}^{\mathcal{A}} = \frac{1}{n}\sum_{x=1}^{n}\max_j u_j(x).
\label{eq:oracle}
\end{equation}
This is an empirical upper bound over the sampled grid for the decomposable objective in Equation~\eqref{eq:deployment}. Although unavailable at inference because it uses ground truth, it quantifies recoverable threshold headroom. It is not an upper bound for non-decomposable corpus metrics.

\noindent\textbf{Computational Characteristics} Caching $Y$ avoids repeated base-model inference. Generic annotation evaluates $k$ choices per unit in $O(nkC_{\mathcal{P},m})$ time and stores $O(nk)$ values. Binary masks admit cumulative counts: once score maps are available, constructing all 257-point direct-utility profiles for the 800 Kvasir training/validation images takes 0.58 seconds and 0.82 MB of raw storage in one Intel Xeon 8255C run (7.79 seconds including loading and feature extraction). With $B=128$, the joint target contains 256 scalars.

\noindent\textbf{Deployment} At deployment, one pass of $\mathcal{M}$ produces $Y$ and $F$. The head predicts $(\hat{\mathbf{a}},\hat{\mathbf{b}})$, and Equation~\eqref{eq:retarget} returns the chosen metric's threshold in place of a fixed binarization value. No ground truth, joint target, or online utility replay is required.

%% file: sec/5_experiments.tex
\section{Experiments}
\label{sec:experiments}

\subsection{Experimental Protocols}
\label{sec:protocols}

We ask three questions: (1) does joint confusion prediction transfer better than scalar objectives and simple adaptive rules; (2) can one checkpoint retarget precision--recall preferences; and (3) does the same decoder generalize to a corruption family excluded from policy training? Every reported utility decomposes per image as required by Equation~\eqref{eq:deployment}.

For mask binarization, we compare fixed 0.5, a validation-selected global threshold, and per-image Otsu~\citep{otsu1979threshold}; Otsu uses its continuous output rather than our grid. A ground-truth-selected threshold appears only as the sampled oracle. Target-global diagnostics use external labels and are unavailable to deployed methods. Per-image Dice is $(2TP+\epsilon)/(2TP+FP+FN+\epsilon)$ and IoU is $(TP+\epsilon)/(TP+FP+FN+\epsilon)$, with $\epsilon=10^{-7}$; all reported values are image macro-averages. Intervals use 10,000 paired bootstrap samples. Kvasir intervals resample images within each fixed dataset because patient/video identifiers are unavailable. KITTI intervals cluster all conditions of each original image. Within every declared split or outer fold, fitting and selection exclude the reported test images.

The supplementary material provides exact runners, manifests, tests, result JSON, and full environment details; runs use Python 3.10/PyTorch 2.6 on a Tesla V100 32 GB GPU.

\subsection{Cross-Dataset Polyp Segmentation}
\label{sec:polyp}

\noindent\textbf{Protocol} We first study image-level mask binarization on Kvasir-SEG~\citep{jha2020kvasir}. Images are resized to $352\times352$ and split into 700/100/200 training/validation/test images; detailed tables use the seed-20260726 reference split. We train an ImageNet-initialized ResNet34 U-Net~\citep{he2016deep,ronneberger2015unet} with BCE and Dice losses using AdamW (batch 16, learning rate $10^{-4}$, weight decay $10^{-4}$), independent horizontal and vertical flips with probability 0.5, and at most 50 epochs with validation patience 8. After selecting its checkpoint by validation Dice, we freeze it. Only the 700 training tuples fit threshold heads; the 100 validation tuples select the global threshold, head checkpoint, and objective.

Each score map yields 61 label-free features: mean, standard deviation, minimum, maximum, binary entropy, fractions above 0.5 and above the map mean; 13 quantiles at $\{.001,.01,.025,.05,.1,.25,.5,.75,.9,.95,.975,.99,.999\}$; a normalized 32-bin histogram on $[0,1]$; and nine foreground fractions at thresholds 0.1 through 0.9. AutoThreshold uses $B=128$ and a 61-96-48-256 ReLU MLP with dropout 0.10/0.05 (23,152 parameters), versus 23,201 for the 257-value direct utility head. Kvasir validation selects $\lambda_{ab}=\lambda_q=.01$. The capacity-matched scalar baseline is a 61-183-64-1 MLP with 23,187 parameters; the original scalar trunk has 10,657. For tied scalar labels, ascending-grid argmax selects the smallest threshold. All neural heads use AdamW (batch 128, learning rate $10^{-3}$, weight decay $10^{-4}$), at most 800 epochs, and validation patience 100. Three fixed head seeds quantify conditional variability for the reference segmenter and split.

We then freeze the complete Kvasir-trained pipeline and transfer it to CVC-300, CVC-ClinicDB, CVC-ColonDB, and ETIS-LaribPolypDB under the standard polyp segmentation protocol~\citep{fan2020pranet}. No images, labels, or thresholds from these four datasets enter training or model selection.

\begin{table*}[t]
\centering
\small
\setlength{\tabcolsep}{3.5pt}
\begin{tabular}{lccccccc}
\toprule
Dataset & N & Fixed .5 & Global & Otsu & Target global$^{\dagger}$ & Joint & Oracle \\
\midrule
\multicolumn{8}{l}{\textit{Held-out source-domain test}}\\
Kvasir-SEG & 200 & \underline{0.8903} & 0.8902 & \textbf{0.8912} & 0.8904 & 0.8885 & 0.9238 \\
\midrule
\multicolumn{8}{l}{\textit{External target datasets without target tuning}}\\
CVC-300 & 60 & 0.7873 & 0.7876 & \underline{0.7994} & 0.7883 & \textbf{0.8106} & 0.8632 \\
CVC-ClinicDB & 62 & 0.6566 & 0.6537 & \underline{0.6603} & 0.6655 & \textbf{0.6691} & 0.7698 \\
CVC-ColonDB & 380 & 0.5583 & 0.5584 & \underline{0.5826} & 0.5595 & \textbf{0.5837} & 0.6764 \\
ETIS-LaribPolypDB & 196 & 0.4325 & 0.4324 & \underline{0.4533} & 0.4329 & \textbf{0.4662} & 0.5511 \\
\bottomrule
\end{tabular}
\caption{Cross-dataset Dice for the reference Kvasir pipeline. Joint averages three head seeds; Otsu uses its continuous threshold. Target global$^{\dagger}$ is selected with each target's labels and is diagnostic, not deployable.}
\label{tab:crossdset}
\end{table*}

\noindent\textbf{Cross-Dataset Results} Table~\ref{tab:crossdset} contrasts source-selected policies with a label-informed target-global diagnostic. On Kvasir test, a single threshold already suffices. Under transfer, Joint is the strongest deployable method listed for every target dataset and reaches $0.6324 \pm 0.0027$ equal-domain macro Dice, versus 0.6080 for Global and 0.6239 for continuous Otsu. Its paired gain over Otsu is $+0.0085$ $[+0.0028,+0.0150]$. Pooled over images instead of domains, Joint scores 0.5778 versus 0.5718, a gain of $+0.0060$ $[-0.0006,+0.0122]$. The effect is therefore positive under both aggregations, but its interval excludes zero only for equal-domain averaging.

\noindent\textbf{Objective and Structure Controls} Table~\ref{tab:controls} holds score maps, features, source split, and validation protocol fixed. Matching the scalar head's capacity does not improve transfer; hard labels, soft utility classification, and expected-regret training also remain below Otsu. Source validation selects soft-target temperature from $\{.005,.01,.02,.05,.1\}$ and tunes the Ridge and GBRT searches. Replacing scalar targets with direct utility profiles raises external Dice by 0.0242 for the MLP and 0.0328 for Ridge, showing that the supervision target, rather than parameter count, drives the gain. Joint is $+0.0023$ over direct utility MLP $[-0.0017,+0.0065]$; direct utility Ridge is $+0.0018$ over Joint $[-0.0112,+0.0073]$. Both intervals span zero. Area-only, scalar Ridge, and scalar GBRT stay near Global, arguing against a simple foreground-ratio explanation.

\begin{table}[t]
\centering
\small
\setlength{\tabcolsep}{1.5pt}
\begin{tabular}{lcc}
\toprule
Method & Kvasir test & External \\
\midrule
Fixed 0.5 & 0.8903 & 0.6087 \\
Global threshold & 0.8902 & 0.6080 \\
Otsu~\citep{otsu1979threshold} & \textbf{0.8912} & 0.6239 \\
Area-only linear (scalar) & \underline{0.8904} & 0.6087 \\
Ridge (scalar, 61 features) & 0.8879 & 0.6014 \\
GBRT (scalar, 61 features) & 0.8899 & 0.6073 \\
MLP scalar regression & 0.8880$\pm$0.0003 & 0.6059$\pm$0.0104 \\
Capacity-matched scalar & 0.8878$\pm$0.0005 & 0.5997$\pm$0.0023 \\
MLP grid classification & 0.8868$\pm$0.0028 & 0.6034$\pm$0.0061 \\
Soft utility classification & 0.8900$\pm$0.0001 & 0.6126$\pm$0.0015 \\
Expected-regret policy & 0.8903$\pm$0.0000 & 0.6100$\pm$0.0019 \\
Direct utility MLP & 0.8888$\pm$0.0008 & 0.6301$\pm$0.0020 \\
Direct utility Ridge & 0.8890 & \textbf{0.6342} \\
Joint confusion (ours) & 0.8885$\pm$0.0000 & \underline{0.6324$\pm$0.0027} \\
Oracle & 0.9238 & 0.7151 \\
\bottomrule
\end{tabular}
\caption{Objective and structure controls on frozen reference outputs. ``External'' averages four target datasets equally. Neural entries report mean $\pm$ sample standard deviation over three head seeds.}
\label{tab:controls}
\end{table}

\begin{figure*}[t]
  \centering
  \includegraphics[width=\textwidth]{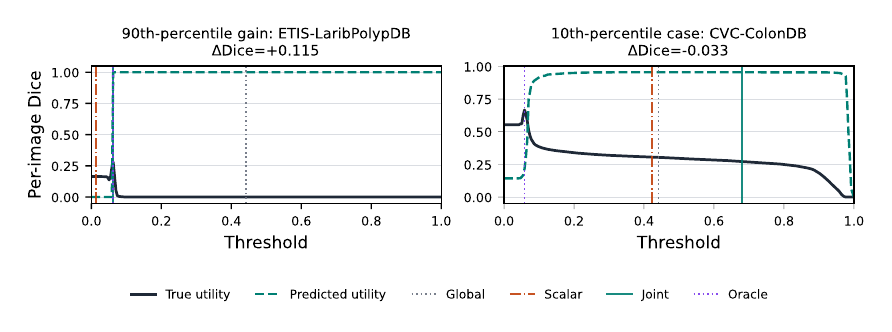}
  \caption{True and predicted Dice profiles for representative external images. Vertical lines mark thresholds from source Global, scalar regression, Joint, and the sampled Oracle. The profile supplies a maximizer rather than a calibrated utility estimate; the right case exposes a remaining selection error.}
  \label{fig:profiles}
\end{figure*}

\noindent\textbf{Decision Geometry and Metric Retargeting} At $\epsilon=.005$, the near-optimal set occupies 0.144 of the threshold range on average, and 11.7\% of external profiles contain multiple disconnected near-optimal components. Joint-minus-Scalar gain correlates negatively with near-optimal width (Spearman $\rho=-.294$): mean gains from the narrowest to widest quartile are $+.1015,+.0187,-.0027,+.0005$. Thus Joint helps most when scalar threshold error is costly. Mean external decision regret drops from 0.1174 for scalar regression to 0.0878 for both direct utility and Joint. Figure~\ref{fig:profiles} visualizes high- and low-gain cases. In Table~\ref{tab:retarget}, changing only $\beta$ in Equation~\eqref{eq:retarget} yields improvements over Otsu for all three utilities.

\begin{table}[t]
\centering
\small
\begin{tabular}{lccc}
\toprule
Method & $F_{0.5}$ & Dice/$F_1$ & $F_2$ \\
\midrule
Global & 0.6250 & 0.6080 & 0.6351 \\
Otsu~\citep{otsu1979threshold} & \underline{0.6305} & \underline{0.6239} & \underline{0.6418} \\
Joint confusion & \textbf{0.6577} & \textbf{0.6324} & \textbf{0.6485} \\
Oracle & 0.7328 & 0.7151 & 0.7483 \\
\bottomrule
\end{tabular}
\caption{Metric retargeting on four external datasets. Each of three seeds selects one checkpoint by Kvasir validation Dice; changing $\beta$ then requires no retraining or target data.}
\label{tab:retarget}
\end{table}

\noindent\textbf{Sensitivity and Profile Error} With one fixed head seed, $B=\{32,64,128,256\}$ gives external Dice $\{.6326,.6300,.6351,.6306\}$, while $\{65,129,257\}$ grid points give $\{.6334,.6341,.6351\}$. For $\lambda_q=\{0,.01,.1,1\}$ results are $\{.6337,.6351,.6271,.6264\}$; over-enforcing marginal reconstruction hurts the decision. On external data, prevalence MAE is .0494, TP-curve MAE .0266, FP-curve MAE .0149, and score-survival MAE .0294. These errors quantify the approximation behind the plug-in decision: the head does not reconstruct every score map exactly.

\noindent\textbf{Observed-Marginal Control} A stricter alternative hardwires the histogram available at inference: $\hat{a}_b = q_b\sigma(r_b)$ and $\hat{b}_b = q_b[1-\sigma(r_b)]$. With $B=128$ and source-selected composition weight .01, this control reaches 0.6095 external Dice. Increasing to $B=256$ matches Joint's 23,152 parameters but reaches 0.6070; both trail continuous Otsu at 0.6239 and Joint at 0.6351 for the same fixed head seed. Thus exact score-marginal reconstruction is not sufficient for accurate decisions, supporting soft marginal regularization rather than a hard constraint. The supplement includes full per-dataset outputs.

\noindent\textbf{Full-Pipeline Variability} For seeds 20260727 and 20260728, we regenerate the split, retrain and select the segmenter, cache new source and external score maps, and refit the heads without external selection. Across all three pipelines, Joint-minus-Otsu differences are $+.0112,+.0027,+.0101$: a mean $+0.0080\pm0.0046$, positive in 3/3 runs. Direct utility is likewise positive in 3/3, by $+0.0055\pm0.0017$. Joint exceeds direct utility in 2/3 pipelines. The transfer gain therefore survives complete-pipeline variation, although Joint does not uniformly dominate direct value learning.

\begin{table}[t]
\centering
\small
\setlength{\tabcolsep}{3pt}
\begin{tabular}{lcccc}
\toprule
Split/model seed & Global & Otsu & Direct utility & Joint \\
\midrule
20260726 & 0.6080 & 0.6239 & \underline{0.6279} & \textbf{0.6351} \\
20260727 & 0.5941 & 0.6078 & \textbf{0.6130} & \underline{0.6105} \\
20260728 & 0.6286 & 0.6458 & \underline{0.6531} & \textbf{0.6558} \\
\bottomrule
\end{tabular}
\caption{Complete Kvasir pipeline repeats. Each row uses a new split and newly trained segmenter; external Dice averages four datasets equally. The head seed is fixed.}
\label{tab:repeats}
\end{table}

\subsection{Cross-Corruption Threshold Adaptation}
\label{sec:corruption}

\noindent\textbf{Setup} We next test detachment from the base model and generalization across corruptions on KITTI Road~\citep{fritsch2013new}. We freeze NVIDIA's SegFormer-B0 Cityscapes checkpoint at $512\times1024$ resolution\footnote{\url{https://huggingface.co/nvidia/segformer-b0-finetuned-cityscapes-512-1024}}~\citep{xie2021segformer,cordts2016cityscapes} and use its road-class softmax probability as the score map; the segmenter is never fine-tuned on KITTI. Because official test annotations are private, we evaluate the 289 labeled training images by fivefold nested cross-validation with seed 2027, stratified by the three KITTI Road categories. Every image occurs in exactly one outer test fold.

The clean experiment uses a compact 61-32-16-1 scalar MLP with 2,529 parameters and the same explicitly defined score-map statistics as Kvasir. It regresses the per-image Dice-optimal threshold on a 101-point grid. Ties favor the threshold nearest the fold's global anchor, then the smaller threshold. Inner validation selects both checkpoint and shrinkage $\alpha\in\{0,.05,\dots,1\}$, where $\hat{t} = t_g + \alpha(\mathcal{N}(F)-t_g)$ and $t_g$ is the validation-tuned global threshold.

The corruption experiment compares this scalar policy with direct utility regression and the joint decoder in Equation~\eqref{eq:decode}. All share the nested LOCO splits, 61-32-16 trunk, optimizer, and validation-Dice criterion. The utility MLP has 101 sigmoid outputs and 4,229 parameters; Joint uses $B=50$, 100 softmax outputs, $\lambda_{ab}=\lambda_q=.01$, and 4,212 parameters. Both profile heads select a 101-point predicted argmax without scalar-label tie breaking or shrinkage. All KITTI heads use AdamW (learning rate $10^{-3}$, weight decay $10^{-3}$), at most 600 epochs, and validation patience 100. Selection batches are 32 for clean and 128 for corruption evaluation; final refits use batch 32 on the complete outer training fold for the validation-selected epoch count. Dice and IoU are image macro-averages excluding official invalid pixels. We designed this experiment after observing the initial clean-transfer result, so it is developmental rather than an untouched confirmation benchmark; nested cross-validation nevertheless prevents fold-level leakage.

\begin{table}[t]
\centering
\small
\setlength{\tabcolsep}{4pt}
\begin{tabular}{lcc}
\toprule
Method & Dice & IoU \\
\midrule
\multicolumn{3}{l}{\textit{Clean images}}\\
Fixed 0.5 & 0.8795 & 0.7985 \\
Otsu~\citep{otsu1979threshold} & 0.8790 & 0.7977 \\
Validation-tuned global & \underline{0.9067} & \underline{0.8376} \\
Scalar regression & \textbf{0.9072} & \textbf{0.8398} \\
Per-image oracle & 0.9142 & 0.8512 \\
\midrule
\multicolumn{3}{l}{\textit{Leave-one-corruption-family-out (LOCO)}}\\
Clean-validation global & 0.8735 & 0.7958 \\
Otsu~\citep{otsu1979threshold} & 0.8721 & 0.7874 \\
Seen-family global & 0.8767 & 0.7988 \\
Scalar regression & 0.8856 & 0.8106 \\
Direct utility profile & \textbf{0.8961} & \textbf{0.8237} \\
Joint confusion profile & \underline{0.8939} & \underline{0.8213} \\
Held-family-aware global$^{\dagger}$ & 0.8920 & 0.8168 \\
Per-image oracle & 0.9087 & 0.8423 \\
\bottomrule
\end{tabular}
\caption{KITTI Road grouped out-of-fold results. LOCO excludes the evaluated corruption family from both policy training and selection. The best non-oracle result per setting is bold.}
\label{tab:kitti}
\end{table}

\noindent\textbf{Clean and Corruption Results} On clean images (Table~\ref{tab:kitti}), scalar regression reaches 0.9072 Dice and 0.8398 IoU, versus 0.9067 and 0.8376 for the inner-validation-tuned global threshold. The paired Dice difference is +0.0006 with a 95\% bootstrap interval of $[-0.0017,+0.0027]$: clean performance is comparable, and the data do not establish an improvement.

Motivated by common-corruption evaluation~\citep{hendrycks2019benchmarking}, we construct brightness (factors .75/.50), gamma darkening ($\gamma=1.4/2.0$), contrast (factors .70/.40), white veil ($\alpha=.20/.40$), Gaussian blur ($\sigma=1.2/2.5$), and JPEG (quality 50/20). Every variant of an original image stays in the same outer fold. In each of six LOCO runs, both severities of one family are absent from inner training and validation. Clean images and the other five families select the head and epoch count; scalar-only anchor and shrinkage selection use the same seen-family validation units. Evaluation uses only outer-test images under the held family. The held-family-aware global in Table~\ref{tab:kitti} is diagnostic because its validation labels reveal that family.

Under LOCO, Joint reaches 0.8939 Dice: +0.0082 over scalar regression [+0.0064,+0.0101] and +0.0218 over Otsu [+0.0179,+0.0258]. Its gain over scalar is positive in all six held families, and it exceeds the label-informed held-family global by 0.0019. Direct utility remains strongest at 0.8961, with a paired +0.0022 advantage over Joint [+0.0007,+0.0038]. Intervals use a paired cluster bootstrap over original-image identity. Joint marginal L1 error is 0.0447 and prevalence MAE is 0.0247. KITTI thus provides complementary evidence for profile rather than scalar supervision, while also showing that decision-consistent direct utility can be better for a fixed metric.

\begin{table}[t]
\centering
\footnotesize
\setlength{\tabcolsep}{1.5pt}
\begin{tabular}{lcccccc}
\toprule
Family & Scalar & Direct & Joint & Aware$^{\dagger}$ & J$-$S & J$-$D \\
\midrule
Brightness & 0.9029 & \textbf{0.9098} & \underline{0.9076} & 0.9073 & +0.0048 & $-$0.0022 \\
Gamma & 0.8949 & \textbf{0.9005} & \underline{0.8979} & 0.8969 & +0.0029 & $-$0.0026 \\
Gaussian blur & 0.8140 & \textbf{0.8532} & \underline{0.8493} & 0.8492 & +0.0353 & $-$0.0039 \\
JPEG & 0.9016 & \underline{0.9029} & \textbf{0.9030} & 0.8944 & +0.0014 & +0.0001 \\
Low contrast & 0.9013 & \textbf{0.9058} & \underline{0.9034} & 0.9031 & +0.0021 & $-$0.0024 \\
White veil & 0.8991 & \textbf{0.9043} & \underline{0.9021} & 0.9010 & +0.0030 & $-$0.0023 \\
\bottomrule
\end{tabular}
\caption{LOCO Dice by excluded family. Aware uses labels from that family and is diagnostic. J$-$S and J$-$D compare Joint with scalar and direct utility.}
\label{tab:loco}
\end{table}

%% file: sec/6_conclusion.tex
\section{Discussion and Conclusion}
\label{sec:conclusion}

AutoThreshold is a post-hoc threshold policy, not a replacement segmenter. Its core contribution is a threshold-sufficient joint target: positive and negative score mass yield coherent confusion curves and multiple $\Fb$ decisions. Joint prediction avoids products of conditional mean prevalence and class-conditional rates; marginal reconstruction remains soft.

The representation exposes auditable invariants. Defining $\widehat{TN}_j = \sum_b \hat{b}_b - \widehat{FP}_j$ yields a nonnegative, unit-sum confusion vector. Each score bin contributes to every lower threshold, coupling neighboring decisions through cumulative mass. These geometric, not accuracy, guarantees distinguish Joint from scalar labels and unconstrained utilities.

The experiments identify when adaptation matters. One global threshold suffices on source Kvasir and clean KITTI. Under cross-dataset transfer, Joint exceeds Otsu in all three pipelines and helps most at narrow utility optima. Its reference-pipeline difference from direct utility Ridge has an interval spanning zero; its distinctive benefit is retraining-free metric retargeting, not uniformly higher Dice. Oracle headroom therefore diagnoses when adaptation is worthwhile.

Developmental KITTI provides a complementary boundary case: LOCO removes the evaluated corruption family from every outer split, and Joint improves over scalar regression but trails direct utility. Because these synthetic transformations informed method development, the evidence supports generalization across predefined families, not arbitrary real-world shift or independent confirmation.

Equation~\eqref{eq:deployment} marks a methodological boundary. Image-, patch-, and dense-level policies require operator and utility to decompose at the same level; corpus ranking metrics and coupled operators such as NMS require groupwise decisions. Within this scope, AutoThreshold offers an auditable route from frozen score maps to adaptive masks while exposing the trade-off between structured plug-in estimation and direct value learning.